\documentclass{article}
\usepackage{spconf,amsmath,graphicx,hyperref}
\usepackage{acronym}
\acrodef{sota}[SOTA]{state-of-the-art}
\acrodef{qa}[QA]{question answering}
\acrodef{tsqa}[TSQA]{time-series question answering}
\acrodef{dnn}[DNN]{deep neural network}
\acrodef{vlm}[VLM]{vision-language model}
\acrodef{llm}[LLM]{large language model}
\newcommand{\numtrials}{five}
\usepackage{amssymb}
\usepackage{multirow}
\usepackage{makecell}
\usepackage{bm}
\usepackage{booktabs}
\usepackage{comment}
\usepackage{pifont}
\usepackage{color}
\usepackage{siunitx}
\usepackage{amsfonts}
\usepackage{adjustbox}
\usepackage{amsmath}
\usepackage{graphicx}
\usepackage{url}
\usepackage{tabularx}
\usepackage{xcolor}

\usepackage[backend=biber,style=ieee,
maxbibnames=6,
doi=false,isbn=false,url=false,eprint=false
]{biblatex}
\defbibheading{bibliography}[\refname]{}

\DeclareSourcemap{
	\maps[datatype=bibtex, overwrite=true]{
		\map{
		    \step[fieldsource=booktitle,
			match=\regexp{(?i).*EUSIPCO.*},
			replace={Proc. EUSIPCO}]
			\step[fieldsource=booktitle,
			match=\regexp{(?i).*CVPR.*},
			replace={Proc. CVPR}]
			\step[fieldsource=booktitle,
			match=\regexp{(?i).*Interspeech.*},
			replace={Proc. Interspeech}]
			\step[fieldsource=booktitle,
			match=\regexp{(?i).*ICASSP.*},
			replace={Proc. ICASSP}]
			\step[fieldsource=booktitle,
			match=\regexp{(?i).*ICLR.*},
			replace={Proc. ICLR}]
			\step[fieldsource=booktitle,
			match=\regexp{(?i).*International.*Joint.*Conference.*on.*Artificial.*Intelligence.*},
			replace={Proc. IJCAI}]
			\step[fieldsource=booktitle,
			match=\regexp{(?i).*IJCNN.*},
			replace={Proc. IJCNN}]
			\step[fieldsource=booktitle,
			match=\regexp{(?i).*ICML.*},
			replace={Proc. ICML}]
			\step[fieldsource=booktitle,
			match=\regexp{(?i).*ASRU.*},
			replace={Proc. ASRU}]
			\step[fieldsource=booktitle,
			match=\regexp{(?i).*SLT.*},
			replace={Proc. SLT}]
			\step[fieldsource=booktitle,
			match=\regexp{(?i).*SSW.*},
			replace={Proc. SSW}]
			\step[fieldsource=booktitle,
			match=\regexp{(?i).*WASPAA.*},
			replace={Proc. WASPAA}]
			\step[fieldsource=booktitle,
			match=\regexp{(?i).*DCASE.*},
			replace={Proc. DCASE}]
            \step[fieldsource=booktitle,
            match=\regexp{(?i).*Detection.*and.*Classification.*of.*Acoustic.*Scenes.*and.*Events.*Workshop.*},
			replace={Proc. DCASE}]
            \step[fieldsource=booktitle,
			match=\regexp{(?i).*MLSP.*},
			replace={Proc. MLSP}]
            \step[fieldsource=booktitle,
			match=\regexp{(?i).*ECCV.*},
			replace={Proc. ECCV}]
			\step[fieldsource=booktitle,
			match=\regexp{(?i).*AAAI.*},
			replace={Proc. AAAI}]
            \step[fieldsource=booktitle,
            match=\regexp{(?i).*Association.*for.*Computational.*Linguistics.*},
            replace={Proc. ACL}]
			\step[fieldsource=journal,
			match=\regexp{(?i)advances.*in.*neural.*information.*processing.*systems.*},
			replace={Advances in NeurIPS}]
            \step[fieldsource=journal,
			match=\regexp{(?i).*TASLP.*},
			replace={IEEE/ACM TASLP}]
            \step[fieldsource=journal,
			match=\regexp{(?i).*JAIR.*},
			replace={JAIR}]
            \step[fieldsource=journal,
			match=\regexp{(?i).*J-STSP.*},
			replace={IEEE J-STSP}]
			\step[fieldsource=journal,
			match=\regexp{(?i).*Transactions.*on.*Knowledge.*and.*Data.*Engineering.*},
			replace={IEEE TKDE}]
			\step[fieldsource=booktitle,
			match=\regexp{(?i).*International.*Conference.*on.*Internet-of-Things.*Design.*and.*Implementation.*},
			replace={Proc. IoTDI}]
			\step[fieldsource=series,
			match=\regexp{.+},
			replace={{}}]
			\step[fieldsource=editor,
			match=\regexp{.+},
			replace={{}}]
			\step[fieldsource=publisher,
			match=\regexp{.+},
			replace={{}}]
			\step[fieldsource=month,
			match=\regexp{.+},
			replace={{}}]
			\step[fieldsource=location,
			match=\regexp{.+},
			replace={{}}]
			\step[fieldsource=address,
			match=\regexp{.+},
			replace={{}}]
			\step[fieldsource=organization,
			match=\regexp{.+},
			replace={{}}]
		}
	}
}

\newcommand{\gray}[1]{\textcolor{gray}{#1}}

\newcommand{\PercentSI}[1]{\SI[number-unit-product={}]{#1}{\percent}}

\title{Can VLM Pseudo-Labels Train a Time-Series QA Model\\That Outperforms the VLM?}
\name{Takuya Fujimura$^{1, 2}$, Kota Dohi~$^{1}$, Natsuo Yamashita$^{1}$, Yohei Kawaguchi~$^{1}$}
\address{$^{1}$ R\&D Group, Hitachi Ltd., $^{2}$ Nagoya University}

\begin{document}
%
\maketitle
\begin{abstract}
Time-series question answering (TSQA) tasks face significant challenges due to the lack of labeled data.
Alternatively, with recent advancements in large-scale models, vision-language models (VLMs) have demonstrated the potential to analyze time-series signals in a zero-shot manner.
In this paper, we propose a training approach that uses pseudo labels generated by a VLM.
Although VLMs can produce incorrect labels, TSQA models can still be effectively trained based on the property that deep neural networks are inherently robust to such noisy labels.
Our experimental results demonstrate that TSQA models are not only successfully trained with pseudo labels, but also surpass the performance of the VLM itself by leveraging a large amount of unlabeled data.
\end{abstract}

\acresetall  

\section{Introduction}
\label{sec:intro}
Time series analysis plays an important role in various domains, such as finance, traffic, and weather~\cite{wu2021autoformer,UCRArchive2018,chow2024towards,kong2025time}.
In particular, the demand for \ac{tsqa} models has been increasing, as these models enable users to ask questions about time series data in natural language~\cite{chow2024towards,kong2025time}.
Also, we aim to develop a domain-independent \ac{tsqa} model unlike previous domain-dependent \ac{tsqa} models~\cite{kong2025time,wang2025itformer,oh2023ecg,xing2021deepsqa,xie2024chatts}.
For example, instead of outputting domain-specific information such as ``the temperature is rising,'' a domain-independent model should output information such as ``the signal has an increasing trend''~\cite{dohi2025domain,ito2024clasp}.
Such domain-independent models can generalize well to novel domains.

One major challenge in developing such a \ac{tsqa} model is the scarcity of labeled data.
First, compared to image and speech datasets, time-series datasets are very limited~\cite{wen2020time}.
Moreover, most general time-series datasets are designed for domain-dependent applications~\cite{UCRArchive2018,kong2025time,wang2025itformer,oh2023ecg,xing2021deepsqa}.
Although several datasets provide pairs of a time-series signal and a domain-independent label~\cite{kawaguchi2024sushi,cai2024timeseriesexam,dohi2025domain}, these datasets either generate synthetic signals based on a signal class~\cite{kawaguchi2024sushi,cai2024timeseriesexam} or estimate the signal class from a given time-series signal~\cite{dohi2025domain}, both by using manually designed functions.
While this approach enables us to construct accurate datasets, the manual design of such functions requires expert knowledge and imposes substantial costs for adding new signal classes.
Thus, the scalability of these datasets is still limited.

Although labeled datasets remain limited, in recent years, \acp{llm} have made great advancements and demonstrated potential for time-series analysis in a zero-shot manner~\cite{xue2023promptcast,liu2023large,kong2025time,gruver2023large,cai2024timeseriesexam,merrill2024language,chow2024towards}.
Several studies have explored the capabilities of \acp{llm} for time-series forecasting~\cite{xue2023promptcast} and QA tasks~\cite{liu2023large,kong2025time,gruver2023large}, where time-series signals are provided as textual inputs.
Furthermore, it has been shown that \acp{vlm}, which receive time-series signals as images, can effectively capture global features and outperform text-based \acp{llm}~\cite{chow2024towards,cai2024timeseriesexam,merrill2024language}.
In addition, \acp{vlm} approach human-level performance when provided with higher-resolution images~\cite{chow2024towards}.
Although \acp{llm} and \acp{vlm} do not always provide accurate information, utilizing them is a promising way.

In this paper, we propose a training approach that utilizes pseudo labels generated by a \ac{vlm}.
To address the scarcity of domain-independent labeled data, we use a \ac{vlm} to generate pseudo labels through natural language interactions, rather than manually designing specific signal-processing-based functions.
Although \acp{vlm} can generate incorrect labels unlike accurate signal-processing-based approaches, we demonstrate that \ac{tsqa} models can still be effectively trained with these pseudo labels, based on the property that \acp{dnn} are generally robust to such noisy labels~\cite{rolnick2017deep}.
Our contributions are follows:
(i) we propose a training framework for \ac{tsqa} tasks that utilizes pseudo labels generated by a \ac{vlm};
(ii) we show that a \ac{tsqa} model trained with pseudo labels outperforms the \ac{vlm} itself by utilizing a large amount of unlabeled data;
(iii) we analyze the impact of noisy labels on the performance of the \ac{tsqa} model; and
(iv) we investigate error patterns of the \ac{vlm}.

\section{Related Work: Training with noisy labels}
\label{sec:training_noisy_labels}
Supervised training requires labeled data.
Although labels are generally assumed to be carefully annotated, datasets sometimes include incorrect labels.
To address this problem, training algorithms robust to noisy labels~\cite{liu2020early} and label-cleansing techniques~\cite{brodley1999identifying} have been studied.

In contrast to these techniques, it has also been shown that \acp{dnn} are inherently robust and can be trained even with noisy labels.
Rolnick et al. showed that, during mini-batch training, the gradient contributions from random noisy labels tend to cancel each other out within a mini-batch, while the consistent gradients from correct labels are enhanced~\cite{rolnick2017deep}.
As a result, \acp{dnn} can be successfully trained despite the presence of noisy labels.
In their experiments, they achieved over \PercentSI{90} image classification accuracy even after adding noisy label data at 100 times the size of the original dataset.
Also, Liu et al. demonstrated that \acp{dnn} first learn from the majority of correct labels and only begin to overfit to noisy labels after the gradients from the correct labels have vanished~\cite{liu2020early}.

Although whether \acp{dnn} eventually overfit to noisy labels depends on the presence of a consistent relationship between the input data characteristics and the incorrect labels, it has been shown that \acp{dnn} can still effectively learn from datasets with noisy labels.
Also, although pseudo-labeling and self-training have been widely used to scale supervision from imperfect teachers~\cite{lee2013pseudo,berthelot2019mixmatch,tarvainen2017mean}, our focus is TSQA: we probe when VLM-generated labels are ``good enough'' and when their systematic errors are inherited.

\vspace{-6pt}
\section{Proposed Method}
\begin{figure}[t]
    \centering
    \includegraphics[width=0.95\linewidth]{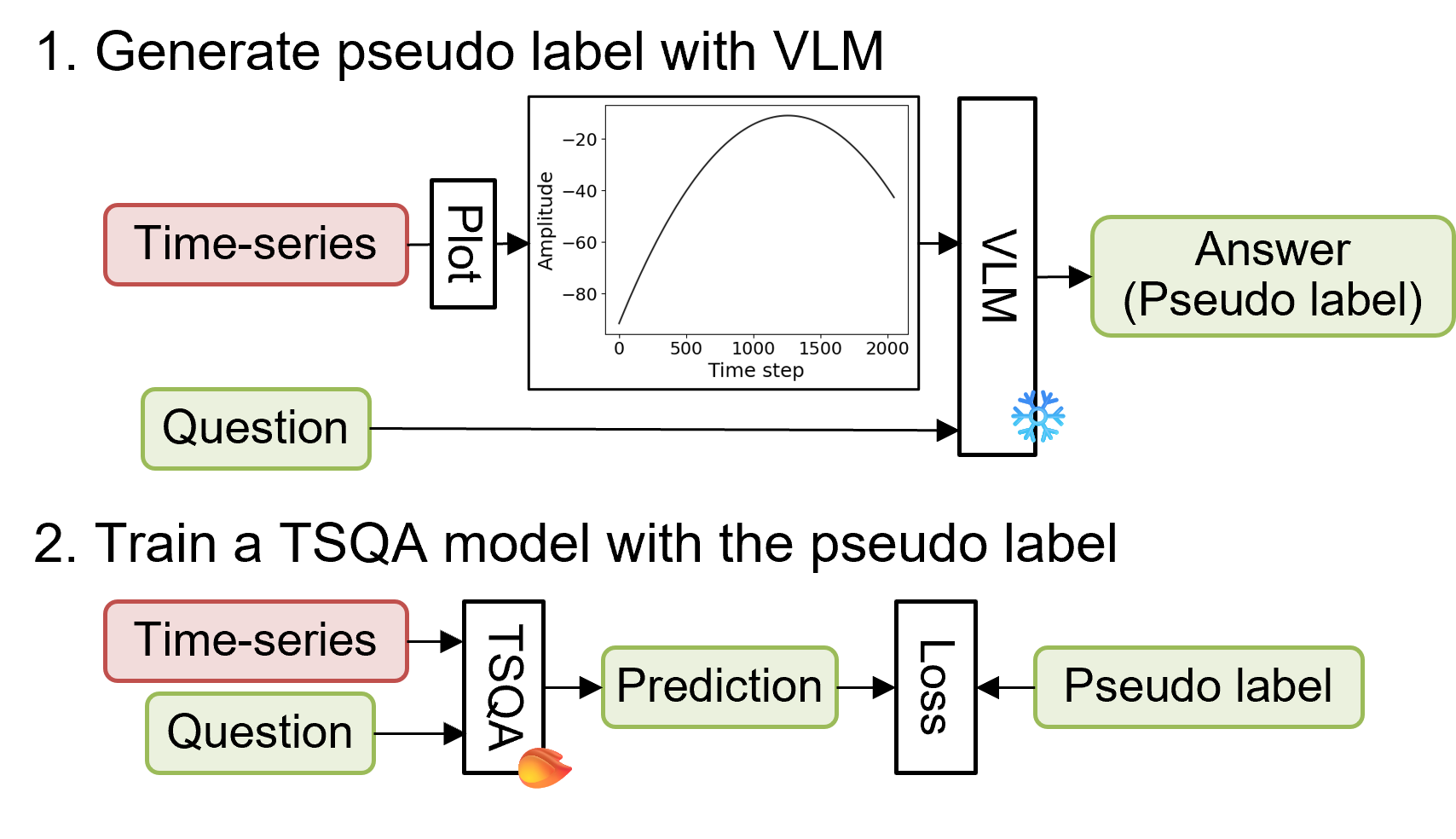}
    \vspace{-5pt}
    \caption{Overview of the proposed method.}
    \vspace{0pt}
    \label{fig:overview}
\end{figure}

\vspace{-5pt}
To construct a domain-independent \ac{tsqa} model without labeled data, we propose to train the model using pseudo labels obtained from a \ac{vlm}.
The proposed method works as follows (Fig.~\ref{fig:overview}).
First, we convert a time-series signal into a plot image (e.g., using \textit{matplotlib}).
Then, we obtain the pseudo label for the time-series signal by inputting the plot image and the question text to the \ac{vlm}.
Finally, we train a \ac{tsqa} model to predict the corresponding pseudo label.
We expect that a \ac{vlm} can provide pseudo labels of sufficient quality for the training.
Also, as discussed in Sec.~\ref{sec:training_noisy_labels}, it is possible to train the model successfully even if the pseudo labels are noisy, provided a sufficient amount of correct labels.

\vspace{-5pt}
\section{Experimental evaluation}
\vspace{-5pt}
We evaluate the effectiveness of our proposed method on a multiple-choice QA task, in which models are required to predict the signal class given a time-series signal and a set of answer options.
Note that the proposed method can also be applied to other tasks (e.g., free-form QA); however, in this study, we focus on the multiple-choice QA task to enable objective evaluation.
We conduct three types of experiments:

\noindent
\textbf{Proof of concept}: We first demonstrate that a \ac{tsqa} model can be trained with pseudo labels generated by \ac{vlm}.

\noindent
\textbf{Requirements for training data}: We conduct simulation experiments to examine the acceptable ratio of incorrect labels and the necessary training data size.

\noindent
\textbf{Analysis of misclassification patterns in pseudo labels}: We analyze the misclassification patterns in the pseudo labels generated by a \ac{vlm}, since the impact of noisy labels also depends on whether consistent error patterns are present.

\vspace{-5pt}
\subsection{Setups}
\vspace{-5pt}
We conducted experimental evaluation using the SUSHI dataset~\cite{kawaguchi2024sushi}, which contains various synthetic time-series signals with domain-independent signal class labels.
Each signal has a length of 2,048 points.
For our experiments, we used clean subsets from the following ten basic classes: \textit{constant (const.)}, \textit{linear increase (lin. inc.)}, \textit{linear decrease (lin. dec.)}, \textit{concave}, \textit{convex}, \textit{exponential growth (exp. growth)}, \textit{exponential decay (exp. decay)}, \textit{sigmoid}, \textit{cubic function (cubic func.)}, and \textit{gaussian (gauss.)}.
The dataset was divided into training, validation, and test sets in a 90:5:5 ratio, yielding 9,000 training samples, 500 validation samples, and 500 test samples.
Each split contained an equal number of samples from each class.

Our \ac{tsqa} model consisted of an \ac{llm} with a time-series encoder, following the previous study~\cite{chow2024towards}.
The time-series encoder extracted an embedding from a time-series signal.
This embedding was concatenated with the text embeddings of the prompt, and the entire sequence was then fed into the \ac{llm}.
Specifically, the input of the \ac{llm} was as follows:
\textit{``\textless s\textgreater[INST] Refer to the following time series signal:\textless\textit{time-series embedding}\textgreater Which pattern does this time series represent? (0) constant (1) linear increase ... (9) gaussian [/INST]''.}
For the \ac{llm}, we used \textit{Mistral-7B-Instruct-v0.1}\footnote{https://huggingface.co/mistralai/Mistral-7B-Instruct-v0.1}, keeping all parameters frozen.
For the time-series encoder, we used a three-layer Informer encoder~\cite{haoyietal-informer-2021}.
The embeddings extracted by the Informer were subsequently processed by average pooling, followed by a two-layer MLP, resulting in a 4,096-dimensional \ac{llm}-compatible embedding.

We trained the model for 100 epochs using the standard cross entropy loss.
The target text was provided in the format ``(\textit{number})'' and the loss was computed only on the target text tokens, while input tokens were masked out.
The optimizer was AdamW~\cite{adamw2019} and the batch size was 32 (distributed as 8 samples per GPU across 4 GPUs).
The learning rate was set to 0.0001 and adaptively reduced by a factor of 0.5 if the validation accuracy did not improve for 2 consecutive epochs. 
We trained the \ac{tsqa} model for \numtrials\ trials, changing both the dataset split and the model initialization.
We evaluated the model on the epoch with the best validation performance.
We compared our proposed method which uses pseudo labels as the target text (TSQA-PL), with the upper-bound method which uses ground-truth labels (TSQA-GT).

For the \ac{vlm}, we used GPT-4o~\cite{openai2024gpt4o} with a temperature of 0.
We input images of time-series signals provided in the SUSHI dataset, each sized at $8 \times 4$ inches with a resolution of 100~dpi, which is considered sufficient~\cite{chow2024towards}.
For GPT-4o, the prompt was:
\textit{``Refer to the time series signal in the image. Please answer the following question. Your answer must be in the format ``(number)'', with the number enclosed in parentheses. No other text is necessary. Which pattern does this time series represent? (1) linear increase ... (9) gaussian''.}

The answer options were shuffled for each sample.
Also, we confirmed that all answers followed the ``(\textit{number})'' format, with one exception that lacked a number.

\begin{table}[t]
\caption{Evaluation results on both training and test sets. Values are represented as ``mean (standard deviation)'' [\%] across \numtrials\ trials.
GPT-4o performance on the training dataset indicates the quality of pseudo labels.
Note that we assume that ground-truth labels are unavailable.}
\label{tab:eval_results}
\begin{tabular}{l|cc}
\toprule
& Train & Test \\
\midrule
Random (chance)                 & 10.00 & 10.00 \\
GPT-4o (baseline)               & 81.71 & 80.20 \\
TSQA-PL (proposed)              & \textbf{92.41 (1.18)} & \textbf{93.12 (1.41)} \\
\gray{TSQA-GT (upper bound)}    & \gray{\textbf{99.87 (0.11)}} & \gray{\textbf{99.92 (0.10)}} \\
\bottomrule
\end{tabular}
\vspace{-15pt}
\end{table}

\begin{figure}[t]
    \centering
    \includegraphics[width=1.0\linewidth]{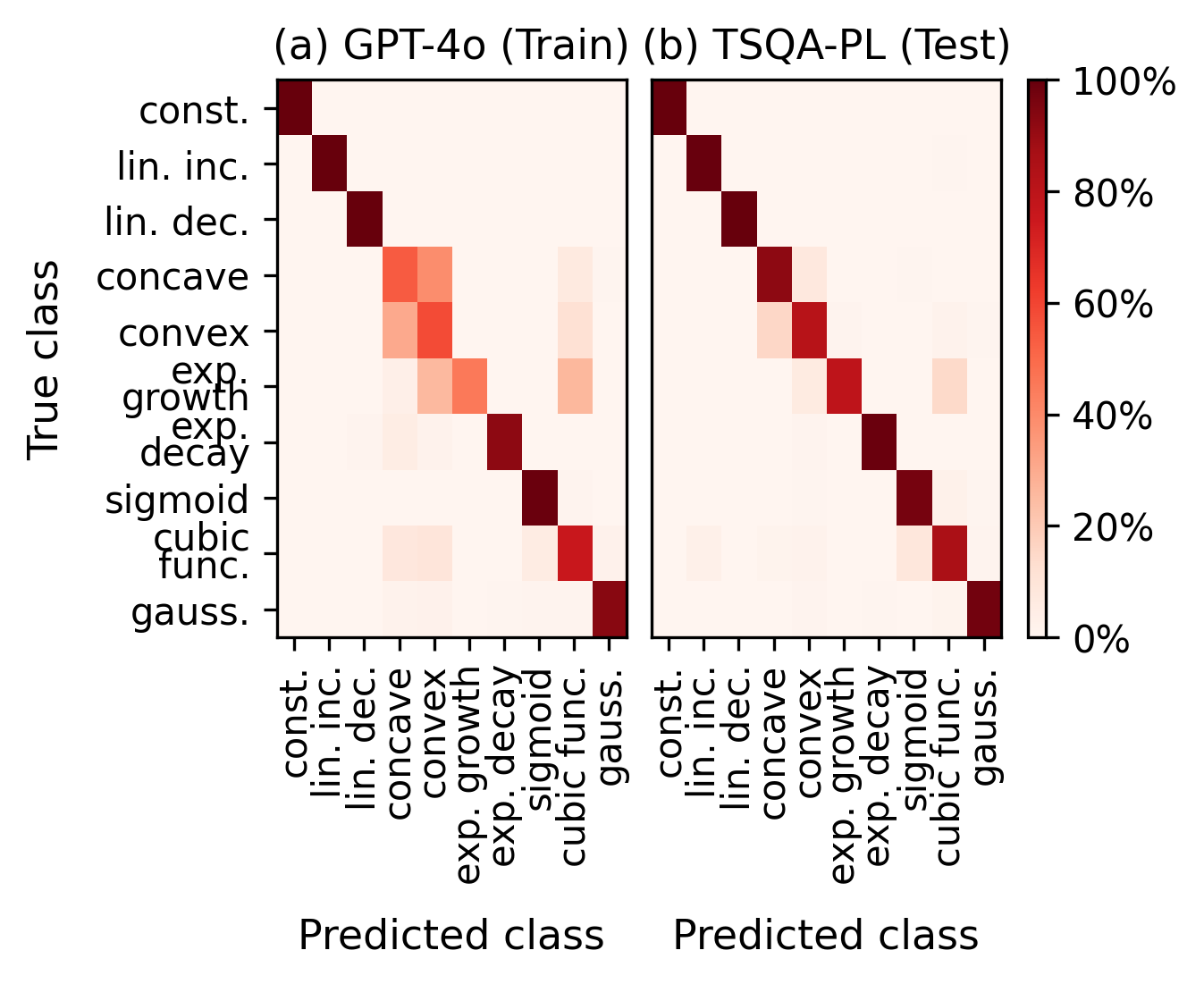}
    \vspace{-15pt}
    \caption{Confusion matrices. (a) Results of GPT-4o on the training set (i.e., pseudo labels used in TSQA-PL) and (b) results of TSQA-PL on the test set, averaged over \numtrials\ trials. The colormap shows the recall score for each class.}
    \label{fig:cm}
    \vspace{-5pt}
\end{figure}

\vspace{-7pt}
\subsection{Proof of concept}
\vspace{-5pt}
Table~\ref{tab:eval_results} shows evaluation results on both the training and test sets.
First, when ground-truth labels are available, the \ac{tsqa} model achieves nearly \PercentSI{100} performance.
Second, GPT-4o demonstrates sufficient performance for pseudo label generation in a zero-shot manner, obtaining correct labels for \PercentSI{81.71} of the training set.
In fact, TSQA-PL is successfully trained and, remarkably, it even surpasses the performance of GPT-4o.
Also, the fact that TSQA-PL outperforms GPT-4o on the training set indicates that TSQA-PL does not overfit to the noisy labels during the training.
Figure~\ref{fig:cm} shows the confusion matrices for GPT-4o and TSQA-PL.
Although TSQA-PL inherits the distribution of pseudo labels produced by GPT-4o, it reduces the errors observed in GPT-4o.

\begin{figure}[t!]
    \centering
    \includegraphics[width=0.96\linewidth]{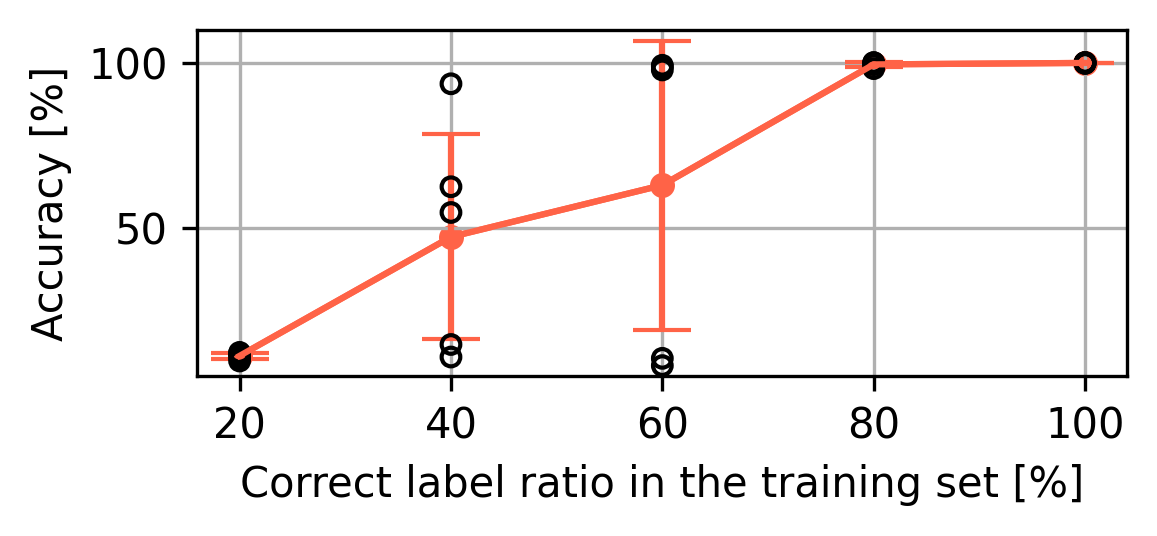}
    \vspace{-15pt}
    \caption{Evaluation results with changing the correct label ratio.
    Black circles represent individual scores from each of the \numtrials\ trials, the red circles represent the mean score, and the red error bars represent the standard deviation.
    }
    \label{fig:acc}
\end{figure}

\begin{figure}[t!]
    \centering
    \includegraphics[width=1.0\linewidth]{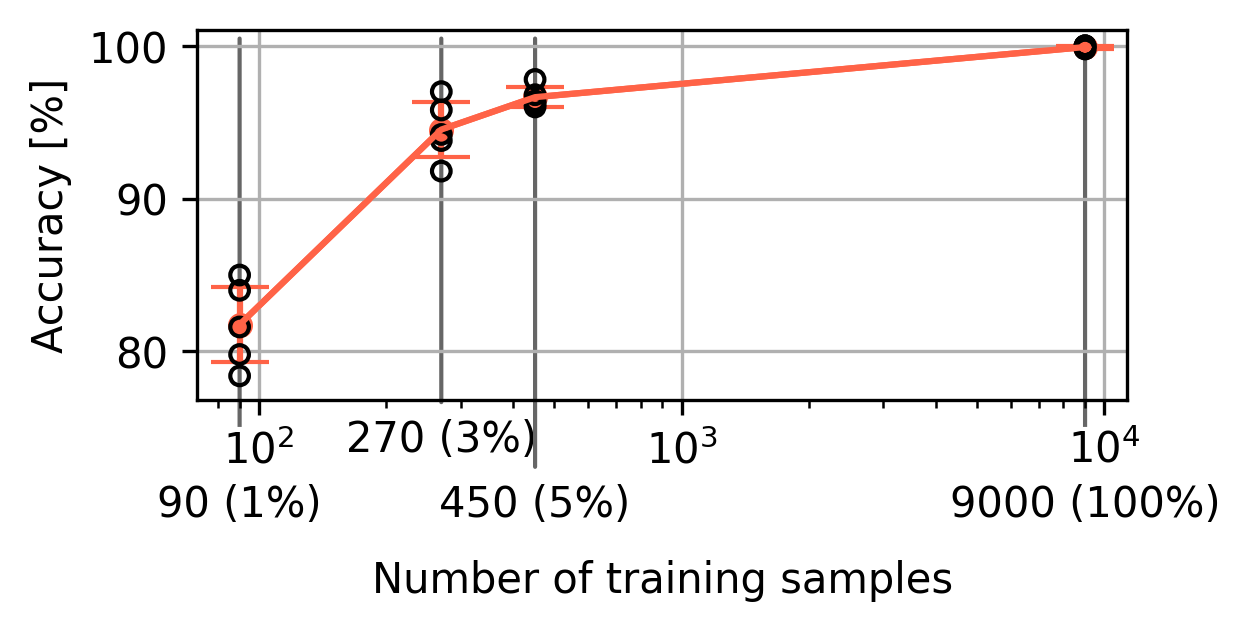}
    \vspace{-25pt}
    \caption{Evaluation results with changing the number of training samples.
    Black circles represent individual scores from each of the \numtrials\ trials, the red circles represent the mean score, and the red error bars represent the standard deviation.
    }
    \label{fig:num_samples}
    \vspace{-5pt}
\end{figure}

\begin{figure*}
    \centering
    \includegraphics[width=0.98\linewidth]{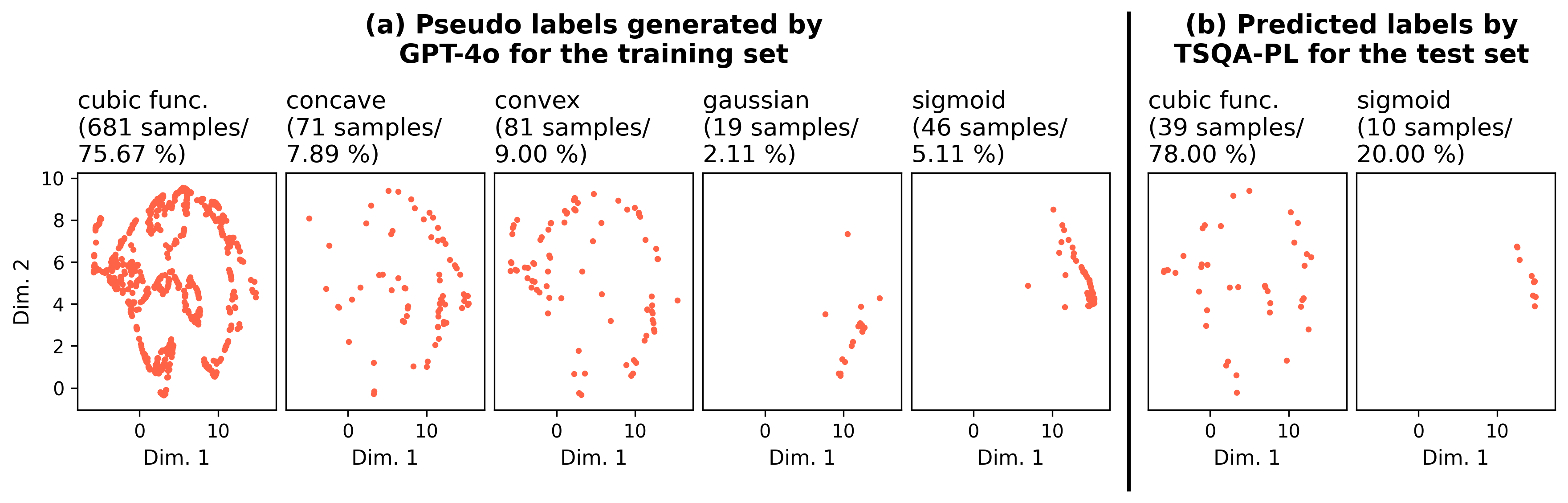}
    \vspace{-12pt}
    \caption{Visualization of the embedding space for the cubic function signals. (a) Embeddings of the training data annotated with pseudo labels generated by GPT-4o, and (b) embeddings of the test data annotated with predictions from TSQA-PL.
    We excluded two samples misclassified as exponential growth in the training set and one sample misclassified as convex in the test set.
    All figures share the same axes.
    These figures show results from a single trial out of \numtrials\ trials.
    }
    \label{fig:embed}
    \vspace{-8pt}
\end{figure*}

\vspace{-13pt}
\subsection{Requirements for training data}
\vspace{-5pt}
To further investigate the above results, we evaluated the performance of the \ac{tsqa} model by changing the correct label ratio, where incorrect labels were randomly selected from the remaining labels excluding the correct label.
The number of training samples was fixed at 9,000.
Figure~\ref{fig:acc} shows the evaluation results.
Although it is evident that the performance degrades with a lower correct label ratio, the model trained with noisy labels still achieves an accuracy higher than the correct label ratio itself.
For instance, when the correct label ratio is \PercentSI{80}, the model achieves a higher accuracy of \PercentSI{99.48}.
Even at a correct label ratio of \PercentSI{40}, it achieves an average accuracy of \PercentSI{47.20} while the variance is large.


In addition, we evaluated the performance by changing the number of training samples while keeping the correct label ratio at \PercentSI{100}. 
Figure~\ref{fig:num_samples} shows the evaluation results.
Although the performance degrades as the number of training samples decreases, the model still achieves an average accuracy of \PercentSI{81.76} even with 90 training samples.
This suggests that a full training set containing 9,000 samples is more than sufficient for the \ac{tsqa} model.
These results indicate that, even when a \ac{vlm} generates incorrect pseudo labels, TSQA-PL can achieve high performance by leveraging a large amount of data, thereby mitigating the negative impact of incorrect labels.

\begin{figure}
    \centering
    \includegraphics[width=0.95\linewidth]{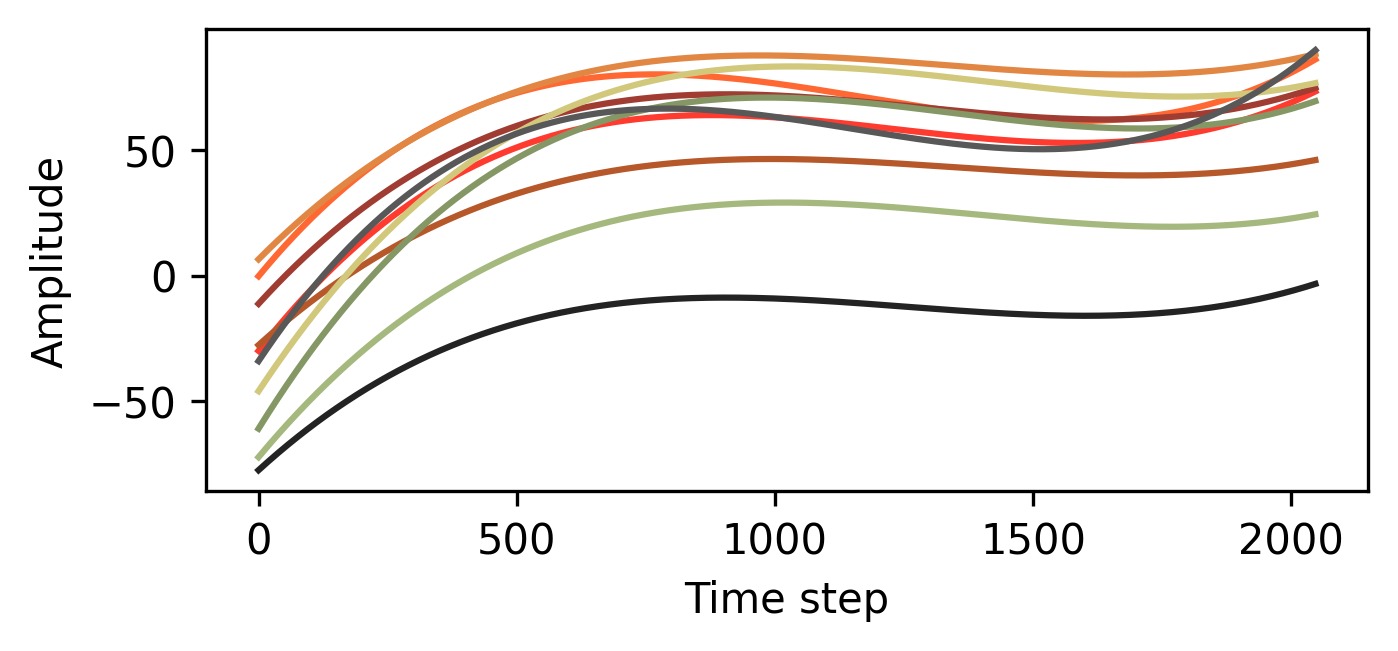}
    \vspace{-15pt}
    \caption{Cubic function signals misclassified as sigmoid by GPT-4o. We randomly selected ten samples for visibility.
    }
    \vspace{-10pt}
    \label{fig:signal}
\end{figure}

\vspace{-11pt}
\subsection{Analysis of misclassification patterns in pseudo labels}
\vspace{-5pt}
We analyze the embedding space of the time-series signals with the labels predicted by GPT-4o (Fig.~\ref{fig:embed}).
We extracted the embeddings from the cubic function signals using TSPulse~\cite{ekambaram2025tspulse} and visualized them with UMAP~\cite{mcinnes2018umap}.
As a preliminary check, we confirmed that TSPulse was able to capture differences in signals as defined by the ground-truth labels.
From Fig.~\ref{fig:embed} (a), we can see that GPT-4o misclassifies some cubic function signals as concave, convex, or gaussian.
However, since these misclassified signals exhibit features similar to those of correctly classified samples, and the majority of such signals are correctly classified, the adverse effect of incorrect labels is mitigated.
On the other hand, GPT-4o incorrectly assign sigmoid labels to most of the signals located in the center-right region of the UMAP plot.
In this case, TSQA-PL learns this relationship and consequently inherits the misclassification as shown in Fig.~\ref{fig:embed}~(b).

Figure~\ref{fig:signal} shows examples of cubic function signals that are misclassified as sigmoid by GPT-4o.
These signals exhibit characteristics distinct from true sigmoid functions, demonstrating the limitations of GPT-4o.

\vspace{-3pt}
\section{Conclusion and Limitation}
\vspace{-3pt}
In this paper, we proposed a training approach that utilizes pseudo labels generated by a \ac{vlm} to address the scarcity of labeled data for \ac{tsqa} tasks.
The proposed method effectively trains \ac{tsqa} models based on the property that \acp{dnn} are generally robust to noisy labels.
Our experimental results demonstrated that
(i) GPT-4o had a sufficient capabilities to generate pseudo labels,
(ii) the \ac{tsqa} model was successfully trained with those pseudo labels,
and (iii) it outperforms GPT-4o itself by utilizing a large amount of unlabeled data.

A limitation of our approach is that the performance depends on the \ac{vlm}.
As shown in Fig.~\ref{fig:embed}, we observed that GPT-4o still exhibits misunderstandings for certain signal characteristics.
Naturally, \acp{vlm} struggle with more complex questions, and, the pseudo labels may not be useful in such cases.
Despite this limitation, we believe our approach remains promising, as the adverse effects of noisy labels can be mitigated by utilizing large amounts of data, and large-scale models continue to improve.
It should also be noted that, although \acp{vlm} struggle with complex questions, obtaining accurate answers for such questions by other approaches is equally costly or difficult.

\section{REFERENCES}
\printbibliography

\end{document}